\documentclass{article}
\usepackage{spconf,amsmath,graphicx,amsfonts,amssymb}
\usepackage[utf8]{inputenc} 
\usepackage[T1]{fontenc}    
\usepackage{hyperref}       
\usepackage{url}            
\usepackage{booktabs}       
\usepackage{amsfonts}       
\usepackage{nicefrac}       
\usepackage{microtype}      
\usepackage{xcolor}

\usepackage{enumitem}
\setlist{nosep, leftmargin=14pt}

\usepackage[compact]{titlesec}

\usepackage{mwe} 

\let\OLDthebibliography\thebibliography
\renewcommand\thebibliography[1]{
  \OLDthebibliography{#1}
  \setlength{\parskip}{0pt}
  \setlength{\itemsep}{0pt plus 0.3ex}
}

\title{On self-supervised multimodal representation learning: \\An application to Alzheimer's disease}
 \name{
    \begin{tabular}{@{}c@{}}
        Alex Fedorov$^{\clubsuit\dagger}$ \qquad
        Lei Wu$^{\diamondsuit\dagger}$ \qquad
        Tristan Sylvain $^{\heartsuit\bullet}$ \qquad
        Margaux Luck$^{\heartsuit}$ \\
        Thomas P. DeRamus $^{\diamondsuit\dagger}$ \qquad
        Dmitry Bleklov \qquad
        Sergey M. Plis $^{\diamondsuit\dagger}$ \qquad
        Vince D. Calhoun $^{\clubsuit\diamondsuit\spadesuit\dagger}$
    \end{tabular}
}
 \address{
    $^{\clubsuit}$ Georgia Institute of Technology,
    $^{\diamondsuit}$ Georgia State University,
    $^{\spadesuit}$ Emory University,\\
    $^{\dagger}$ Center for Translational Research in Neuroimaging and Data Science,
    Atlanta, GA, USA\\
    $^{\heartsuit}$ Mila, ${^\bullet}$ Universit\'e de Montr\'eal, Montr\'eal, Qu\'ebec, Canada
}
\begin{document}
%
\maketitle
\begin{abstract}
Introspection of deep supervised predictive models trained on functional and structural brain imaging may uncover novel markers of Alzheimer's disease (AD). However, supervised training is prone to learning from spurious features (shortcut learning), impairing its value in the discovery process. Deep unsupervised and, recently, contrastive self-supervised approaches, not biased to classification, are better candidates for the task. Their multimodal options specifically offer additional regularization via modality interactions. This paper introduces a way to exhaustively consider multimodal architectures for a contrastive self-supervised fusion of fMRI and MRI of AD patients and controls. We show that this multimodal fusion results in representations that improve the downstream classification results for both modalities. We investigate the fused self-supervised features projected into the brain space and introduce a numerically stable way to do so.
\end{abstract}

\section{Introduction}

Diagnosing pathologies from raw medical imaging outputs is often a
more complex problem than the idealized problems faced in non-medical
image classification. Single modalities often do not contain enough
information requiring multimodal fusion of multiple distinct data sources (as is commonly the case for MRI). Besides, multiple data
sources contain a wealth of complementary information and insufficient
redundancy to align them easily. As a result, correctly utilizing the different sources can be vital to designing robust diagnostic tools.

While supervised approaches might be prone to shortcut learning~\cite{geirhos2020shortcut} and require more data, we tend to explore unsupervised methods~\cite{antiDaubechie2013}.
Some common approaches to tackle learning with multiple sources are inspired by Deep CCA~\cite{andrew2013deep}, parallel ICA~\cite{liu2008parallel}, and recently by variational direction such as MMVAE~\cite{shi2019variational}. A new breed is recently emerging that relies on deep learning in achieving powerful representations using a self-supervised approach.
Specifically, the algorithms are based on the maximization of mutual information (DIM)~\cite{dim}.
However, the machine learning field is breeding a zoo of various seemingly unique methods exploiting this approach~\cite{bachman2019amdim,stdim,sylvain2019locality,tian2019contrastive,chen2020simple} while they can be unified under one paradigm. We investigate all of the existing and generate some yet unpublished methods under the same framework to apply to AD and healthy controls (HC) data on functional (f)MRI and structural (s)MRI modalities. The proposed approach shows great promise empirically.

Our contributions are as follows:
\begin{itemize}
    \item We compare and contrast all approaches by their effect on the downstream classification task.
    \item We also show that representation similarity of the learned AD embeddings does not necessarily lead to better classification yet allows us to uncover links between modalities.
    \item We report an improved and numerically stable method of investigating thus obtained multimodal features via model introspection as a statistical test contrasts.
    \item We performed all experiments on a large AD/MCI dataset.
\end{itemize}

\section{Methods}

\subsection{Problem definition}
Let $\{x^{1}_i, x^{2}_i\}_{i=1}^N$  be the dataset of  N paired images
of  different   modalities:  $1$  is   T1  and  $2$  ---   fALFF  (see
Section~\ref{sec:dataset} for  details).  We want to  learn compressed
and semantically  meaningful latent  representations of  each modality
$\{z^{1}_i, z^{2}_i\}_{i=1}^N$.   The latent  representation $z^*_{i}$
is an vector $\in \mathbb{R}^{d}$ encoding the image $x^*_{i}$ through
encoder  $E^*$  parametrized  by  a  neural  network  with  parameters
$\theta^*$ as $z^*_{i} = E^*(x^*_{i})$.

To learn the set of parameters $\theta^*$ we want to minimize the
objective ${\cal L}$ defined as:
\begin{equation*}
    {\cal L} = -\sum_{i=1}^2 \sum_{j=1}^2 L(M_i,M_j),
\end{equation*}
where $L(M_i,M_j)$  is uni-modal  ($i=j$) or  multi-modal ($i  \ne j$)
objective.   In this  work  we specifically  explore the  decoder-free
objectives based on maximization of mutual information.

\begin{figure}[t]
    \center
    \includegraphics[scale=0.12]{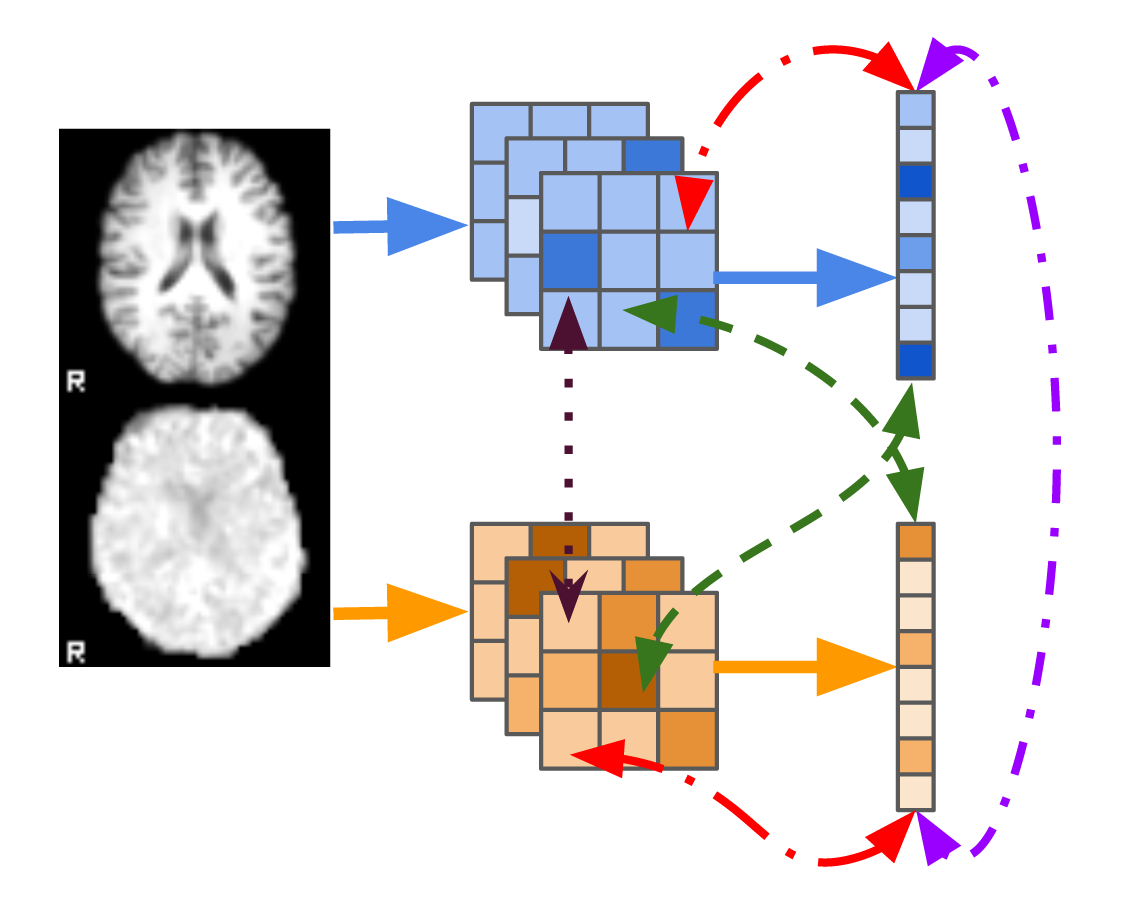}
    \vspace{-0.3cm}
    \caption{The sample pair of T1 and fALFF from OASIS-3 and a
    scheme of the method. First, images are passed through the encoder to get convolutional features and final latent representation, further used by objectives. The arrows represent objectives: red is L, purple --- S, green CL, and magenta CS. For convolutional features, each location is represented as a vector across channels, while latent representation is a whole vector. The arrow connects location/latent with location/latent and has a meaning in the sense of predictability.}
    \label{fig:dataset_scheme}
    \vspace{-0.3cm}
\end{figure}

\subsection{Mutual Information Maximization}
Here  we  utilize the  familiar  InfoNCE~\cite{oord2018representation}
based estimator for a lower bound of mutual information.

\begin{equation*}
    L(M^i,M^j) = I(M^i; M^j) \ge \frac{1}{N} \sum_{m=1}^N \log \frac{e^{f(u^i_m,v^j_m)}}{\frac{1}{N}\sum_{k=1}^N e^{f(u_m^i,v_k^j)}},
\end{equation*}
where $f(u^i_m,v^j_k) = \frac{u^{i\intercal}_m v^j_k}{\sqrt{n}}$ is a separable critic function, where $n$ is a dimension of latent representation. The embeddings $u^i_l$ and $v^j_l$ are computed using additional projections $\psi^i$ or $\phi^j$ parametrized by neural networks for latent representation $z$ or features from $l$th layer $c_l$.

The idea behind this estimator to learn representations such that $f(u^i_l,v^j_l) \gg f(u^i_l,v^j_k)_{l \ne k}$.

\subsection{Constructions of the objectives}

Using these definitions we can construct different ways of maximizing mutual information with multi-source data which are shown schematically in Figure~\ref{fig:dataset_scheme}.
The edges represent pairs of features used in the critic functions.
Specifically, the unimodal objective L (Figure~\ref{fig:dataset_scheme}, red) is known as local deep InfoMax (DIM)~\cite{dim}, where we train $f(\phi^i(c^i_{l,m}), \psi^i(z^i_m))$, where we maximize mutual information betwen convolutional features $c^i_{l,m}$ and latent $z^i_m$.
The simple extension of L to capture information between modalities are cross-local (CL) (Figure~\ref{fig:dataset_scheme}, green) and cross-spatial (CS) (Figure~\ref{fig:dataset_scheme}, magenta) used by AMDIM~\cite{bachman2019amdim}, ST-DIM~\cite{stdim} and CM-DIM~\cite{sylvain2019locality}, where we learn $f(\phi^i(c^i_{l,m}), \psi^j(z^j_k))_{m \ne k}$ and $f(\phi^i(c^i_{l,m}), \phi^j(c^j_{l,k}))_{m \ne k}$ critics, respectively.
Last connection (Figure~\ref{fig:dataset_scheme}, purple) is utilizing latent similarity (S) is originally shown by CMC~\cite{tian2019contrastive} and then perfected by SimCLR~\cite{chen2020simple}, where --- $f(\psi^i(z^i_{m}), \psi^j(z^j_k))_{m \ne k}$.
Each variant implies own inductive bias on predictability between embeddings. As L and CL imply the InfoMax principle.
The objectives of type CS and S maximize the similarity between convolutional features and latent variables on the same level respectively.

\begin{figure*}
  \includegraphics[width=\textwidth]{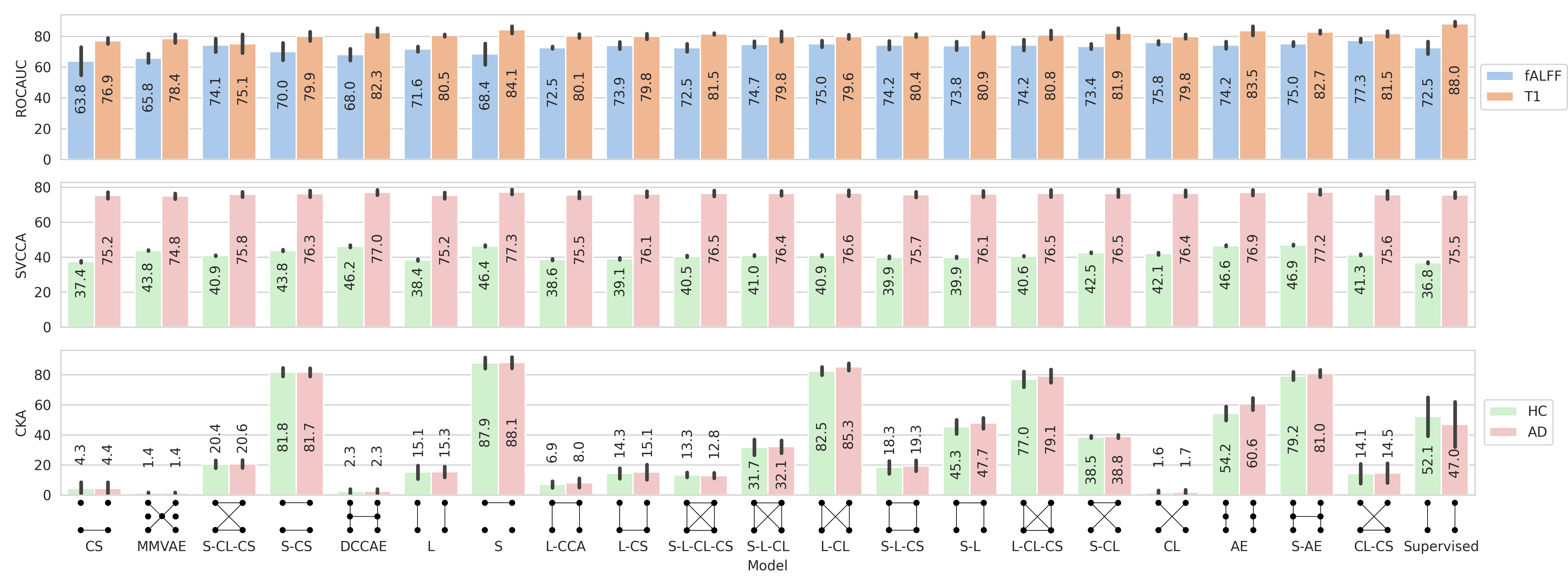}
    \vspace{-1cm}
    \caption{Hold-out  downstream  performance  on OASIS  dataset  with
    Logistic Regression  trained on representations  and cross-modal
    CKA  similarity of the latent representation in different groups. The label letters are L--local, CL--cross-local, CS--cross-spatial, S--similarity. See the main text for details.}
    \label{fig:oasis}
  \end{figure*}

For   completeness,   we   compare  DIM-based   methods   to   related
DCCAE~\cite{wang2015deep},      MMVAE       with      looser      IWAE
estimator~\cite{shi2019variational} ($K=64$).   We also combine the CCA  objective
with L-objective  (L-CCA) and AE  with the Similarity objective  (S-AE) to
create new combinations.   However, for  DCCAE, we do not
pre-train the encoder layer-wise as in the  original work before fine-tuning to a
CCA objective -- our focus here is on the pure end-to-end approaches.
Additionally,  we train a supervised model for an OASIS dataset to get
an approximate bound of
what is achievable to compare to all architectures. The  schemes for
baseline     and    additional     combinations    are     shown    in
Figure~\ref{fig:oasis}.

\section{Experimental setup}

  \subsection{Dataset}
  \label{sec:dataset}
We choose neuroimaging dataset OASIS-3~\cite{LaMontagne2019.12.13.19014902} to study Alzheimer's disease. As modalities, we selected T1 and fractional amplitude of low-frequency fluctuation (fALFF) prepared from T1w and computed from resting-state fMRI, respectively.

Resting-state fMRI time series were registered to the first image in the series using mcflirt in FSL~\cite{fsl} (v 6.0.2), using a 3-stage search level (8mm, 4mm, 4mm), 20mm field-of-view, 256 histogram bins (for matching), 6 degrees-of-freedom (dof) for transformation, a scaling factor of 6mm, and normalized correlation values across images as the cost function (smoothed to 1mm). Final transformations and outputs were interpolated using splines as opposed to the default trilinear interpolation. Then a fractional amplitude of low-frequency fluctuation (fALFF) map was then computed in 0.01 to 0.1 Hz power band using REST~\cite{rest}. After the visual inspection, 15 T1w images were removed. T1w images were brainmasked using bet from FSL. Afterward, both fALFF and T1 images were linearly (7 dof) converted to MNI space and resampled to 3mm resolution. The final volume size for moth modalities is $64\times64\times64$. The data preprocessing is minimized intentionally to reduce its impact on training with deep neural networks and transformed to simplify analysis.

After analyzing demographic data, we only leave Non-Hispanic Caucasians (totaling 826 subjects) since other groups are underrepresented. For the group with Alzheimer's disease (AD), we choose all subjects with confirmed AD records, and the healthy cohort (HC) are cognitively normal subjects. The subjects with other conditions are used as an additional group only in unsupervised pretraining. We combine all possible pairs (4021 pairs) for pretraining, which are closest by days of multimodal images for each subject. During the final evaluation, we leave only one pair for each subject.

We split subjects on 5 stratified (70\% healthy, 15\% AD, 15\% other) cross-validation folds (580-582 subjects (2828-2944 pairs), 144-146 subjects (653-769 pairs)) and hold-out (100 subjects (424 pairs)). Then we apply histogram standardization based on each training subset and z-normalization to images using TorchIO library~\cite{perez_garcia_torchio_2020}. For pretraining on OASIS-3, we use random flips, random crops as data augmentation. During optimization, we also utilize class balanced data sampler~\cite{balanced_data_sampler}.

\subsection{Architecture, Hyperparameters, and Optimization}
In our experiments, we use DCGAN~\cite{radford2015unsupervised}. DCGAN is a convolutional architecture with an encoder and a decoder. The last layer maps input features to a $64$-dimensional latent representation. The convolutional projection heads consist of 2-convolutional layers with kernel size $1$, input equal to the number of features of the selected layer with feature side size $8$ in the encoder, and output --- to $64$. The latent projection heads are chosen to be identity. All the weights of projections are shared across all contrastive objectives.

We penalize each contrastive bound with squared matching scores $\lambda f(u,v)^2$ of the critic with $\lambda = 4\mathrm{e}{-2}$ and clip values of the critic by $c\tanh(\frac{s}{c})$ with $c=20$. The projections are shared across different objectives. Thus, the optimization of the objective can be considered as multi-task learning.

To train the weights of the neural networks, we used
RAdam~\cite{liu2019variance} with learning rate $4\mathrm{e}{-4}$ and
OneCycleLR~\cite{smith2019super} scheduler with maximum learning rate
$0.01$ for $200$ epochs with batch size $64$. However, the model MMVAE
we could train only with batch size $4$ due to memory constraints.

\section{Results}

\subsubsection*{Downstream task}
To evaluate the representation on a downstream task, we trained the Logistic Regression (LR) on top of the representation produced by the pre-trained encoder. To choose hyperparameters of Logistic Regression, we searched the space using Optuna~\cite{optuna_2019} over $5$ fold cross-validation by computing the mean ROC AUC as a score for $500$ iterations.
Inverse regularization strength $C$ is sampled log-uniformly from the  $[1\mathrm{e}{-6}, 1\mathrm{e}{+3}]$ interval, the penalty is chosen from L1, L2, or elastic net, the elastic net mixing parameter is sampled uniformly from unit interval. The solver for LR is saga~\cite{NIPS2014_ede7e2b6}.

The results are shown in Figure~\ref{fig:oasis}. The models are sorted by the average AUC across modalities.
Overall, most combinations of contrastive objectives outperform CCA-based DCCAE and variational MMVAE. The best-unsupervised result for T1 is by unimodal AE $83.5\%$ and multimodal S $84.1\%$, for fALFF --- by multimodal CL-CS $77.3\%$. Comparing fALFF results for S and AE methods, we notice that the performance is lower by $5.8\%$, thus similarity might degrade the performance. Interestingly, while for T1, the supervised model is still the leader by $3.9\%$, the unsupervised method CL-CS surpasses it by $4.8\%$ for fALFF. We argue that the multimodal objective has a regularizing effect. Additionally, the method S-AE might be the right candidate for future analysis. It combines reconstruction error and maximization of mutual information from two perspectives while preserving higher downstream performance and higher similarity of the representation (as we show it further using similarity analysis).

\begin{figure}[ht!]
  \center
  \includegraphics[width=\linewidth]{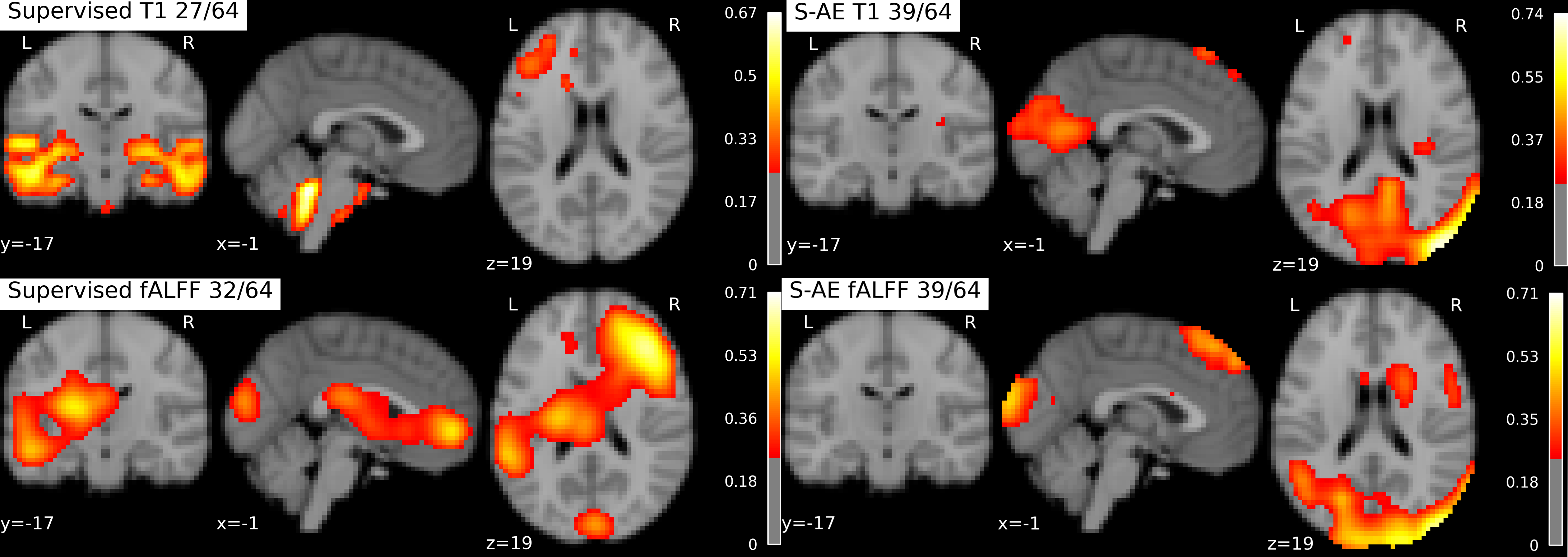}
  \caption{Left column: the highest correlated ($0.29$) pair of mean saliency images for the Supervised method. Right column: the highest correlated ($0.79$) pair of mean saliency images for the S-AE method.}
  \label{fig:correlation_mm}
\end{figure}

\begin{figure}[ht!]
  \center
  \includegraphics[width=\linewidth]{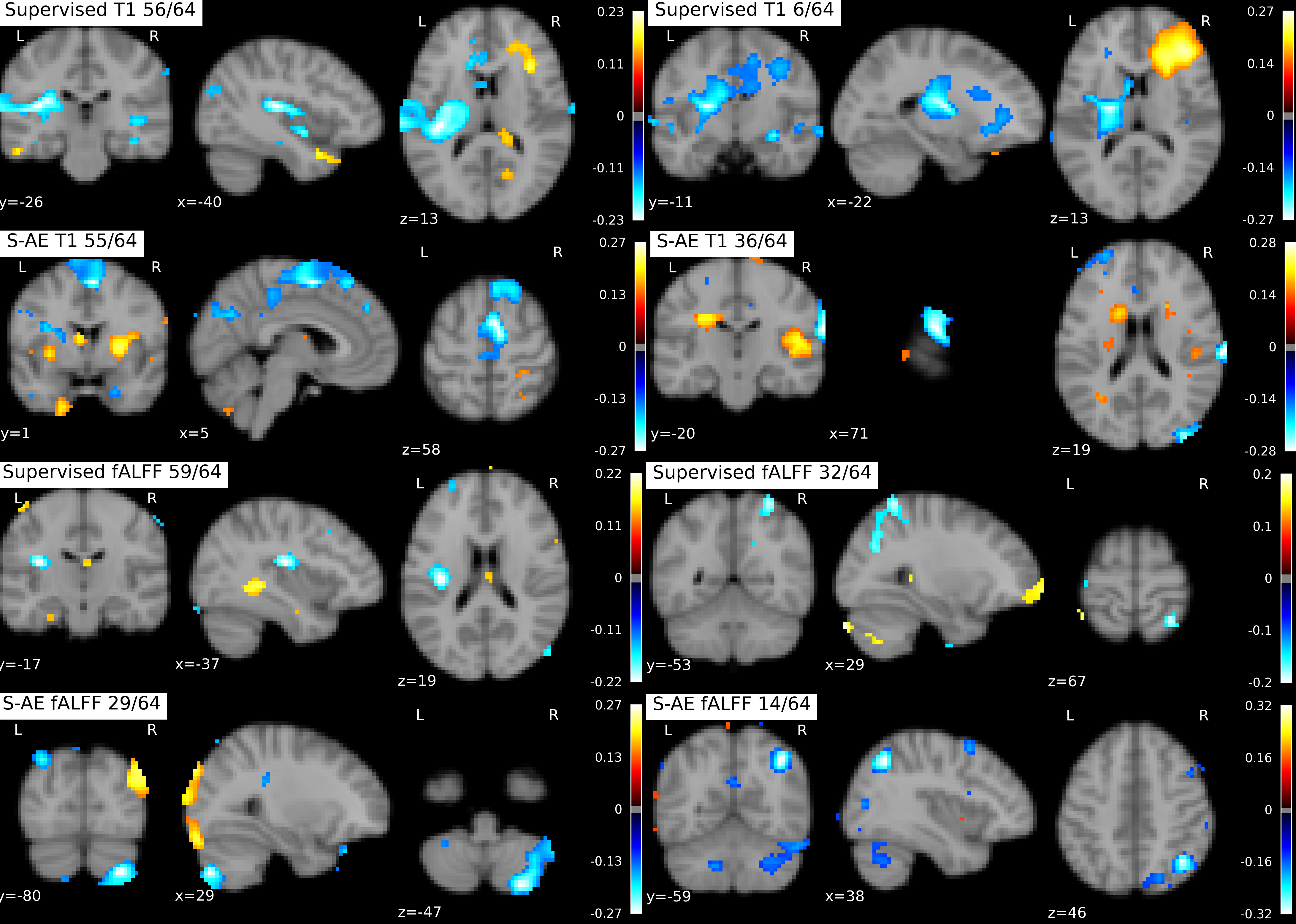}
  \caption{Group differences on T1 and fALFF are shown using effect size RBC. The left column is for dimensions with the highest positive beta in Logistic Regression, right column --- with the highest negative beta. Odd rows are for the Supervised model, even rows --- for S-AE. The coordinates are chosen by the absolute peak value of the effect size.}
  \label{fig:groupdiff}
\end{figure}

\subsubsection*{Representational Insights using similarity}
To better understand the influence of different multi-source objectives on the latent representation, we employ SVCCA~\cite{NIPS2017_7188} and CKA~\cite{kornblith2019similarity}.
We compute the similarity of the representation between modalities in
different groups.
The metrics are shown in Figure~\ref{fig:oasis}.

Per SVCCA metric, models behave similarly across AD patients. At the same time, the AE, S-AE, DCCAE, S, and S-CS have a noticeable difference compared to other models on HC, indicating a higher correlation. The AD patients also have a higher correlation between modalities compared to HC subjects. This might indicate that healthy subjects have richer representation within a modality, thus fewer similarities between modalities. Given performance on the downstream task, we can conclude that unsupervised learning can capture group differences even without prior knowledge about them.

The CKA metric shows significant differences in representation between
models, even though most of the methods are very close in their
predictive performance. This empirically supports the hypothesis that
similarity does not guarantee higher downstream performance or can have a regularizing effect.

\subsubsection*{Representational Insights using saliency}

To gain an additional understanding of how representations behave in uni-source and multi-source, supervised and unsupervised settings with respect to brain and groups, we utilize sensitivity analysis based on SmoothGrad~\cite{smilkov2017smoothgrad} with std $0.05$ and $5$ iterations. We compute gradients for each dimension of the latent representation instead of computing them based on a label. After computing sensitivity maps, we apply brain masking, rescale gradient values to a unit-interval, smoothing them with a Gaussian filter ($\sigma=1.5$).

Using computed saliencies, we study how dimensions of the latent representation correspond to the input image in unimodal and multimodal scenarios. Given one dimension in T1 and another in fALFF, we compute correlations across subjects for each dimension. Then we select the highest correlated pairs ($0.79$ for S-AE and $0.29$ for Supervised) and show thresholded mean saliency on each modality in Figure~\ref{fig:correlation_mm}. Unsupervised multimodal method S-AE shows highly spatially related saliencies between modalities. In contrast, the supervised method has no relation between modalities.

Using Logistic Regression, we selected dimensions with the highest positive and negative beta values. Then we study the group differences using a voxel-wise Mann-Whitney U Test on computed saliencies and report rank-biserial correlation (RBC) as effect size. The results for S-AE and Supervised model are shown in Figure~\ref{fig:groupdiff}. Both methods, Supervised and S-AE, ``look'' at different regions on both modalities. While the supervised method captures regions from labels, they might be trivial markers that might not benefit our understanding of brain degeneration.
Unsupervised methods learn more general representation while being discriminative. For example, S-AE captures non-trivial locations (Figure~\ref{fig:groupdiff}) in the brain, which might be attractive and need to be analyzed much more closely.

\section{Conclusions}
We investigated previous and introduced new approaches for multimodal representation learning using advances in self-supervised learning. Applying our approach to the OASIS dataset, we evaluated learned representation with multiple tools and obtained strong empirical insights for further development in data fusion. Our findings indicate the high potential of DIM based methods for addressing the shortcut learning problem.

\section{Compliance with Ethical Standards}
This research study was conducted retrospectively using human subject data made available in open access by OASIS-3~\cite{LaMontagne2019.12.13.19014902}. Ethical approval was not required, as confirmed by the license attached with the open-access data.

\section{Acknowledgments}
\label{sec:acknowledgments}
This study is supported by NIH R01 EB006841.

Data were provided in part by OASIS-3: Principal Investigators: T. Benzinger, D. Marcus, J. Morris; NIH P50 AG00561, P30 NS09857781, P01 AG026276, P01 AG003991, R01 AG043434, UL1 TR000448, R01 EB009352. AV-45 doses were provided by Avid Radiopharmaceuticals, a wholly-owned subsidiary of Eli Lilly.

\bibliographystyle{IEEEbib}
\bibliography{references}

\end{document}